\documentclass[letterpaper, 10 pt, conference]{ieeeconf}  

\IEEEoverridecommandlockouts                              

\usepackage{amsmath} 
\usepackage{amssymb}

\title{\LARGE \bf
Transformers for Multi-Object Tracking on Point Clouds
}

\author{Felicia Ruppel$^{1,2}$, Florian Faion$^1$, Claudius Gl\"{a}ser$^1$ and Klaus Dietmayer$^2$
\thanks{$^{1}$Robert Bosch GmbH, Corporate Research, 71272 Renningen, Germany, 
        {\tt\small \{firstname.lastname\}@de.bosch.com}}%
\thanks{$^{2}$Institute of Measurement, Control and Microtechnology, Ulm University, Germany,
        {\tt\small \{firstname.lastname\}@uni-ulm.de}}%
}
\usepackage{url}
\usepackage[noadjust]{cite}
\usepackage{graphicx}
\usepackage{subfig}
\usepackage{pgfplots}
\pgfplotsset{
	max space between ticks=30pt,
	try min ticks=3,
	every axis/.style={
		axis y line=left,
		axis x line=bottom,
		axis line style={thick,->,>=latex, shorten >=-.4cm}
	},
	every axis plot/.append style={thick},
	tick style={black, thick}
}
\tikzset{
	semithick/.style={line width=1.1pt},
}
\usepackage{booktabs}
\usepackage{multirow}
\usepackage[colorinlistoftodos,color=pink]{todonotes} 
\usepackage{bm}
\presetkeys{todonotes}{inline}{}

\newcounter{todocounter}

\usepackage{fancyhdr}
\usepackage{ragged2e}

\fancypagestyle{FirstPage}{
	
	\lfoot{\noindent\fontsize{7.5}{8.5}\selectfont\justifying\textcopyright\ 2022 IEEE.  Personal use of this material is permitted.  Permission from IEEE must be obtained for all other uses, in any current or future media, including reprinting/republishing this material for advertising or promotional purposes, creating new collective works, for resale or redistribution to servers or lists, or reuse of any copyrighted component of this work in other works.}
	
}

\begin{document}
\maketitle
\thispagestyle{empty}
\pagestyle{empty}

\begin{abstract}
We present TransMOT, a novel transformer-based end-to-end trainable online tracker and detector for point cloud data. The model utilizes a cross- and a self-attention mechanism and is applicable to lidar data in an automotive context, as well as other data types, such as radar. Both track management and the detection of new tracks are performed by the same transformer decoder module and the tracker state is encoded in feature space. With this approach, we make use of the rich latent space of the detector for tracking rather than relying on low-dimensional bounding boxes. Still, we are able to retain some of the desirable properties of traditional Kalman-filter based approaches, such as an ability to handle sensor input at arbitrary timesteps or to compensate frame skips. This is possible due to a novel module that transforms the track information from one frame to the next on feature-level and thereby fulfills a similar task as the prediction step of a Kalman filter.
Results are presented on the challenging real-world dataset nuScenes, where the proposed model outperforms its Kalman filter-based tracking baseline.
\end{abstract}

\section{INTRODUCTION}
\thispagestyle{FirstPage}
Multi-object tracking (MOT) is an important chain link in the perception, prediction and planning pipeline of an autonomous or automated vehicle. The task comprises keeping track of all relevant surrounding objects over time, where every object should have a unique identity that remains unchanged in the time frame of interest. On a frame-by-frame basis, each track can have four different states. \emph{Initialization:} The object is seen for the first time and its track starts. \emph{Continuation:} The same objects is visible in the following measurement(s), each time appending its new position to the track. \emph{Occlusion:} An object can be occluded for a certain time and possibly reappear in the measurements afterwards. \emph{Completion:} It is seen for the last time at some point, concluding its track. Since a multi-object-tracking module provides the current and past state of the surrounding agents of a vehicle, it is an essential component for the subsequent modules of the driving pipeline such as prediction of the agents' future movement, which, in turn, is essential for planning the behaviour of the ego vehicle. 

There exist multiple fundamentally different approaches to tracking. Classically, the task has been addressed with Bayesian approaches, such as joint
probabilistic data association filters \cite{bar2011tracking} and random finite set based filters \cite{mahler2007statistical}, which are especially suited for measurements with known noise characteristics. With the emergence of reliable deep-learning based object detectors, less computation-intensive Bayesian approaches are widely used for 3D MOT \cite{weng_3d_2019, chiu_probabilistic_2020}: In \cite{weng_3d_2019}, the authors found that their plain Kalman-filter based approach could achieve state-of-the art results on the nuScenes dataset \cite{caesar_nuscenes:_2020}, powered by the output of a reliable object detector, and it outperformed more complicated tracking methods such as \cite{zhang_robust_2019}. The aforementioned tracking approaches can be described as \emph{tracking-by-detection}, which means that tracking is regarded as a capsuled problem, i.e. detections serve as input and tracks are output, commonly on a frame-by-frame basis. Each detection is described by a set of parameters, such as location, extent, and orientation. Approaches that belong to this group can function without access to raw measured data and can be very fast. However, the trackers are restricted to their low-dimensional input, which does not include higher-dimensional features that the detector inferred from the input data. This additional information, however, has the potential to provide useful insights, e.g. for recognition of an object based on its appearance. To mitigate this drawback, some workarounds were proposed, such as utilizing re-identification features \cite{chiu_probabilistic_2020-1}.
\begin{figure}[t]
	\centering
	\includegraphics[width=\columnwidth]{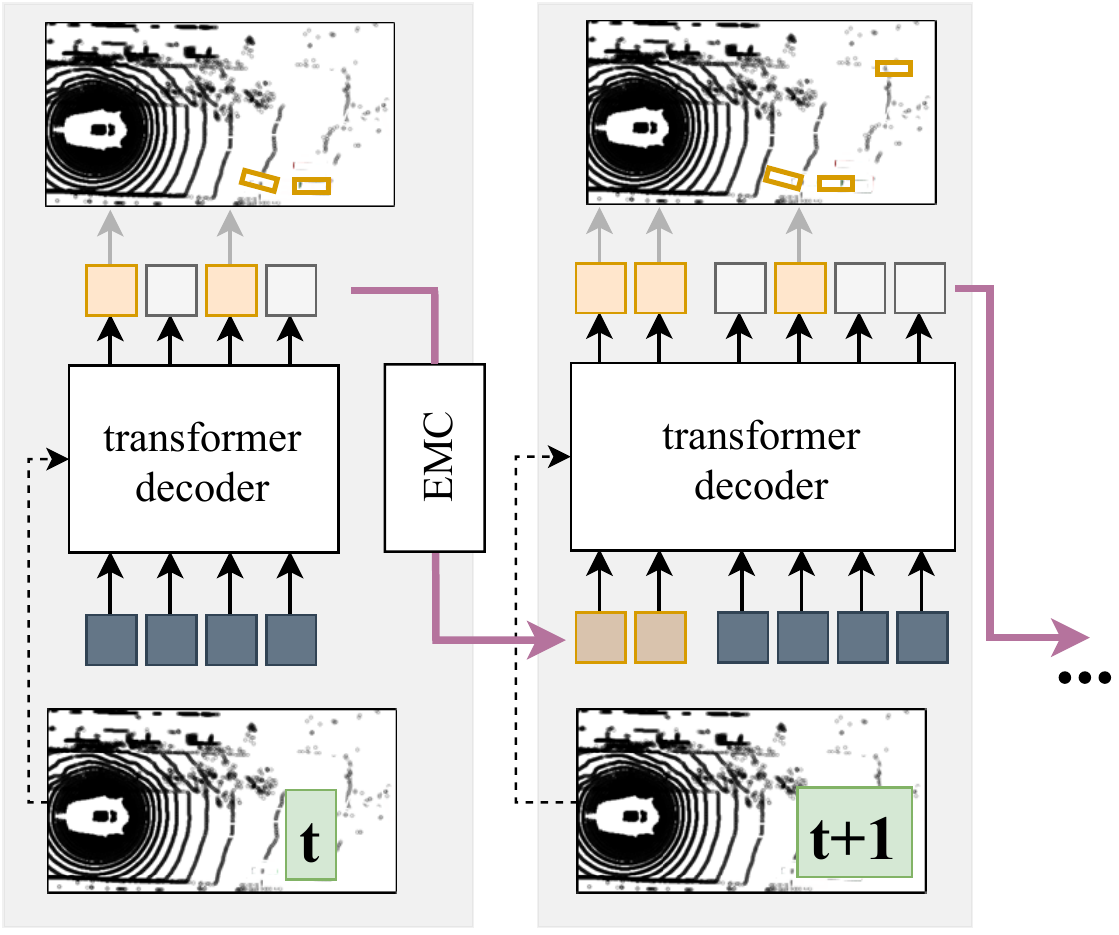}
	
	\caption{Illustration of one tracking step. A main component is our novel ego-motion compensation (EMC) module that transforms the tracker state in feature space. Refer to Fig.~\ref{fig_tracker_large} for a full model overview, as some details are not pictured in this graphic, which is implied by dashed lines.}	\label{figTrSmall}
\end{figure}

In this work, we propose a model, which is illustrated in Figure~\ref{figTrSmall}, that performs tracking and detection jointly to make use of said latent space that an object detector has. This paradigm, \emph{tracking-and-detection}, has recently become more popular, especially in the area of image-based 2D tracking, as its potential became apparent~\cite{meinhardt_trackformer:_2021}. Some existing approaches on lidar data take a stack of frames as input, e.g. the past $N$ frames, and some perform different tasks jointly, such as detection, tracking and prediction \cite{luo_fast_2018}. In our model on the other hand, a sequence is processed on a frame-by-frame basis, avoiding overlap that occurs with stacked input frames.

In 2017, transformers were proposed as a novel machine learning model based on dot-product attention \cite{vaswani_attention_2017}, which revolutionized the field of natural language processing. Besides this, the transformer showed promising results in vision tasks \cite{dosovitskiy_image_2020}, such as object detection \cite{carion_end--end_2020} and tracking \cite{meinhardt_trackformer:_2021} on the image domain and shape classification on lidar data \cite{engel_point_2020}.

Track management requires a global overview of all tracks and object candidates. This makes a transformer a natural fit for this task since it can employ self-attention between the tracklets. However, image data and lidar data behave fundamentally different when it comes to detection and tracking, therefore it is non-trivial to apply existing approaches in the lidar domain. We develop a model that performs both detection and tracking on point cloud data, such as lidar, based on a transformer.

One peculiarity of lidar data in an automotive setting is that the ego vehicle can travel significant distances between two frames, making an ego-motion compensation (EMC) necessary. Because the tracker state is encoded implicitly in feature space in our proposed model, we develop a novel trainable EMC module to perform the compensation in this space. The module can ego-motion correct the tracker state to any point in time, enabling us to be robust against frame skips and delayed sensor input, which is a common occurrence in practice \cite{shin_roarnet_2019}. 

To summarize, the main contributions of this work are:

\begin{itemize}
	\item A transformer-based model for 3D detection is presented that is applicable to point clouds, particularly lidar data in an automotive application, and that serves as a building block for tracking in latent space.
	\item We develop a novel ego-motion compensation (EMC) module that operates in feature space and can be used to compensate skipped frames.
	\item We propose a novel tracking-and-detection model based on a transformer that can accept new input frames at arbitrary time steps. It is modular and therefore suitable for further enhancements such as multi-modality.
\end{itemize}
\vspace{5pt}
\section{RELATED WORK}
This work belongs to the field \textit{tracking-and-detection}, for which related work is detailed this section, as well as the Bayesian baseline that we compare our method to: ProbMOT \cite{chiu_probabilistic_2020}. Since we build upon the transformer model \cite{vaswani_attention_2017}, we outline its functionality.
\subsection{Multi-Object Tracking (MOT)}
\label{sec_related_work_MOT}
ProbMOT \cite{chiu_probabilistic_2020} is a Bayesian \textit{tracking-by-detection} method, where the association problem between existing tracks and new measurements is solved with a Mahalanobis distance-based matching. Then, a Kalman filter update incorporates the new measurements, followed by a prediction for the next frame, where each of the tracked agents is described with a constant-velocity motion model. The tracking model operates within a global coordinate frame, therefore a motion correction of the ego-vehicle is not necessary. ProbMOT performed best at the nuScenes \cite{caesar_nuscenes:_2020} tracking challenge at NeurIPS 2019, which is why we use it as a baseline to compare our approach with a traditional Bayesian \textit{tracking-by-detection} method.

In this paper, we continue a new line of work, denoted as \textit{tracking-and-detection}. For image-based 2D tracking, it is already common to replace parts of the tracking pipeline, such as feature extraction, affinity computation or association with a deep learning model \cite{ciaparrone2020deep}. Furthermore, there exist models that tackle the tracking and detection task with a single model, such as \cite{meinhardt_trackformer:_2021,zhou2020tracking}. In \cite{bergmann2019tracking}, a detector's regression head is used to regress each track to the next timestep. A framework for training deep multi-object trackers is proposed in \cite{xu2020train}. In recent years, \textit{tracking-and-detection} approaches for MOT on point clouds also emerged, aiming to enrich the tracking with detector knowledge. A model-based method is proposed in \cite{fortin2015model}. In \cite{huang2021joint}, an affinity computation and data association module based on deep learning is developed. Some methods combine even more tasks into one model and operate on a stack of consecutive frames \cite{luo_fast_2018, liang2020pnpnet}. Different from other methods, our approach performs in a frame-by-frame online manner and does not need a separate data association model, as track association is performed implicitly by the transformer. Besides this, it is designed to not be a 'black-box' deep learning model, but rather an extendable approach that performs track association and measurement update, but the transition of the tracker state from one frame to the next can be modeled as desired, for which we introduce a novel EMC module.

\subsection{Transformers and Attention}
Our method is based on a transformer model \cite{vaswani_attention_2017}, which is briefly introduced in the following. For a more thorough review, the reader is referred to \cite{vaswani_attention_2017}. A main principle that transformers are based on is \textit{attention}. Input to the model is a sequence of tokens, where every token is a feature vector $\bm{x}_i, i=1,\dots,N$.

For \textit{self-attention} between the tokens, a query $\bm{q}_i$, key $\bm{k}_i$ and value $\bm{v}_i$ vector is computed from each token by multiplying it with the learnt weight matrices $\bm{W}_Q$, $\bm{W}_K$ and $\bm{W}_V$, respectively. The attention weight of a token $\bm{x}_a$ to another, $\bm{x}_b$, $a,b\in\{1,\dots ,N\}$, is simply the dot product $\bm{q}_a\cdot \bm{k}_b$. To compute the attention weights for all token-to-token combinations, a compact matrix notation can be used: $\bm{Q}\bm{K}^\top$, stacking the query and key vectors together. This weight matrix is scaled and applied to the stacked values $\bm{V}$ to obtain
\begin{equation}\label{eq_attention}
\textrm{Attention}(\bm{Q}, \bm{K}, \bm{V}) = \textrm{softmax}\left(\frac{\bm{Q}\bm{K}^\top}{\sqrt{d_k}}\right)\bm{V},
\end{equation}
where $d_k$ is the dimension of the keys and queries. For \textit{cross-attention} between two sets of tokens, rather than \textit{self-attention}, the query vectors are computed from the second sequence $\bm{y}_i, i=1,\dots,M$. Besides this, the attention module is unchanged. In the transformer model \cite{vaswani_attention_2017}, multiple attention heads are employed in parallel to allow each of them to focus on different aspects. Besides, multiple multi-head attention blocks are stacked on top of each other to facilitate a larger field of view, i.e. not only one-to-one attention, but many-to-many.
A full transformer consists of an encoder and a decoder. In the encoder, \textit{self-attention} is applied to the input. In the decoder, both kinds of attention can be found: \textit{self-attention} within its (decoder) input tokens as well as \textit{cross-attention} between the decoder tokens and the output tokens of the encoder. What is input to the decoder depends on the task.

Transformers have been applied to object detection on images \cite{carion_end--end_2020}, where the input to the model encoder is a sequence of pixel embeddings on a coarse grid that is computed by a backbone. The decoder input tokens, on the other hand, serve as slots for the detection of objects and are learnt vectors.
\section{PROPOSED MODEL}
\begin{figure*}[!t]
	\vspace{5pt}
	\includegraphics[width=\textwidth]{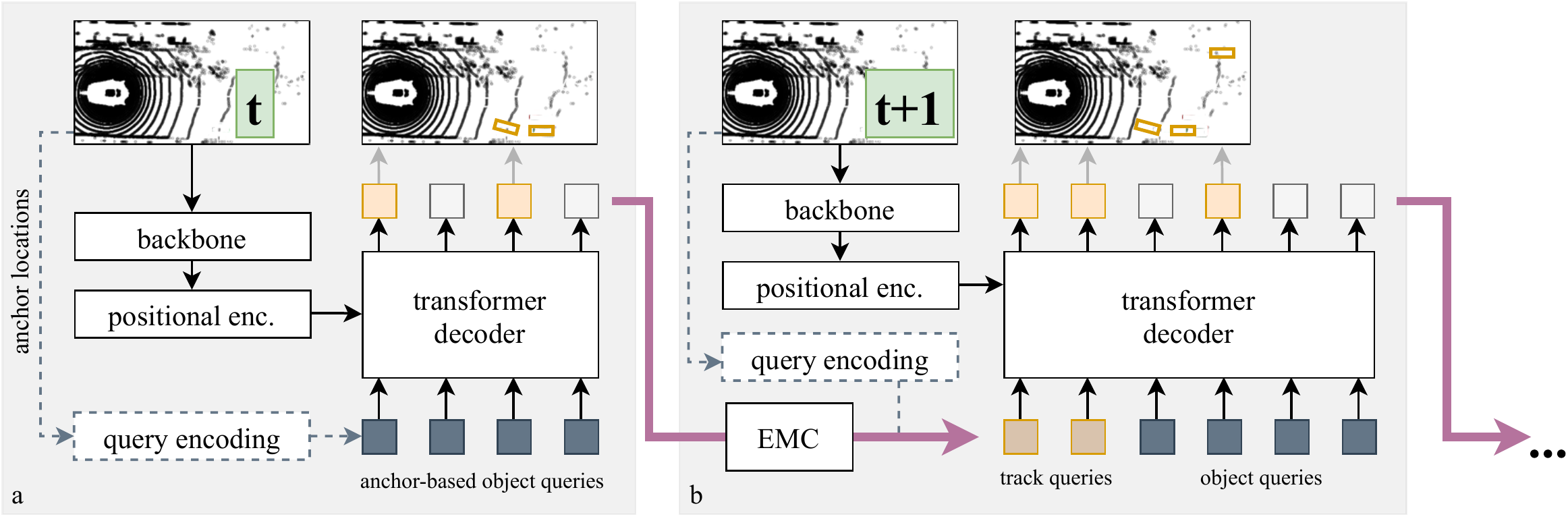}
	\caption{Tracking and detection model overview. a. Detection model, which can function as a standalone detector as well as a track initializor in the first frame of a sequence. The input point cloud is pre-processed through a backbone and a positional encoding is added to retain the location information of the computed feature vectors. Through a sampling method, $M$ anchor locations are obtained from the input point cloud, which are encoded to generate object queries for the transformer decoder that serve as slots for possible objects at the output. b. Joint tracking and detection with track and object queries, whereas the track queries are propagated from the previous time step and ego-motion corrected through the proposed EMC module. When a track query results in a valid object at the output of the decoder, it is considered a continued track, while new tracks originate from object queries.}
	\label{fig_tracker_large}
\end{figure*}

An overview of our proposed joint tracking and detection model can be found in Fig.~\ref{fig_tracker_large}. Each of its components is introduced in the following.
\subsection{Object Detection}\label{sec_detector_model}
To perform object detection on lidar data with a transformer, we follow the general approach of the Detection Transformer (DETR) \cite{carion_end--end_2020}. However, it is non-trivial to apply this image-based model, which consists of an image backbone followed by a full transformer with encoder and decoder, to lidar data in an automotive context. One issue is that in a transformer \textit{encoder}, the size of the self-attention matrix grows quadratically with the input sequence length $N$, limiting it to a maximum length of $N_{\textrm{max}}$, which can be as low as about $2000$ tokens, depending on the available GPU type. This is why DETR \cite{carion_end--end_2020} operates on a grid of only $35 \times 45$ cells as input to the encoder, which is feasible for image data. With automotive lidar data on the other hand, objects tend to be small compared to the sensor range. When a grid-based backbone for lidar data, such as PointPillars \cite{lang_pointpillars:_2019} is used before the encoder, this limitation in turn limits the side length of the grid to $\sqrt{N_{\textrm{max}}}$ and therefore bloats the size of each cell. The same holds for a point-based backbone such as PointNet++ \cite{qi_pointnet++_2017}, where the influence region around each subsampled point grows as the number of tokens becomes limited.

An overview of our transformer detector for point clouds is pictured in Fig.~\ref{fig_tracker_large} (a). We found that the backbone we selected, PointPillars \cite{lang_pointpillars:_2019}, can fulfill some of the context encoding that the transformer encoder normally would. Therefore, it is possible to remove said encoder entirely. An alternative to this would be to use one of the modified transformer models that aim to remedy the memory limitation issue that self-attention has \cite{wang_linformer:_2020} \cite{tay_efficient_2020}. Our experiments showed, however, that skipping the encoder entirely yields the best results, because the other remedies still limit the grid size. Following \cite{lang_pointpillars:_2019}, the input to the PointPillars backbone is a point cloud, which consists of three-dimensional coordinates. Within PointPillars, it is discretized into a set of pillars, which are located on a grid on the birds-eye-view plane. The backbone outputs one feature vector per grid cell. Note that PointPillars could be replaced with any other backbone, which is able to output feature vectors that are bound to a certain location in 3D space or on a grid. From this location, a sine and cosine positional encoding \cite{carion_end--end_2020} is computed and added to the vectors. Then, the backbone result is flattened into a sequence of feature vectors and input into the transformer decoder to constitute keys and values for cross-attention. This decoder consists of six layers with eight attention heads. The dimension of the keys, values and queries is $d=256$.
In our model, the queries for the decoder are data-dependent rather than learnt. Inspired by \cite{misra_end--end_2021}, we use a farthest point sampling \cite{qi_pointnet++_2017} to obtain locations $\bm{\rho}_i$ from the data, with $i=1,\dots,M$, and encode them with a Fourier encoding \cite{tancik_fourier_2020, rahimi2007random}:
\begin{equation}\label{eq_query_encoding}
	\bm{y}_i=\textrm{FFN}\left[\sin(\bm{B}\bm{\rho}_i), \cos(\bm{B}\bm{\rho}_i)\right],
\end{equation}
where $\bm{B}\in \mathbb{R}^{\frac{d}{2}\times3}$ is a matrix that contains entries drawn from a normal distribution, the feed-forward network (FFN) consists of two layers with ReLU activation and is trained with the rest of the model and $\bm{Y}=\{\bm{y}_i\}_{i=1}^M$ denotes the set of tokens that is input to the decoder. As introduced in \cite{rahimi2007random}, each line of $\bm{B}$ projects the coordinates $\bm{\rho}_i$ into a different direction, which is chosen randomly in order to approximate a shift-invariant kernel with the inner product of two encoded points. We name the locations $\bm{\rho}_i\in \mathbb{R}^3$ \textit{anchor locations} because the queries computed from them serve as a prior in the search for objects. However, the model is not restricted by them and can find objects at a distance from their anchor. Besides this, the anchor locations are different from anchor boxes, which are commonly used in other detection approaches such as \cite{lang_pointpillars:_2019}.

In our model, the anchor locations change with every frame, since they are drawn from the data in the present input frame. Farthest point sampling is a commonly used, fast strategy to select locations within a point cloud that are spread out well over the data. For lidar data, location-bound queries are superior to learnt queries, because the point cloud can be very sparse. Otherwise, the model would need to waste resources on finding locations that actually contain data.

Now, each decoder query token $\bm{y}_i$, i.e. anchor-based object query, serves as slot for a possible object. From every \textit{output} token $\bm{y}'_i$ of the decoder with anchor $\bm{\rho}_i$, a feed-forward network is used to compute the parameters
\begin{equation}\label{eq_box_params}
	\bm{d}_i = (\Delta x,\Delta y,\Delta z,w,l,h,\gamma, v_x,v_y, {\tt cls}),
\end{equation}
$i=1,\dots , M$, with location $(x,y,z)=\bm{\rho}_i+(\Delta x,\Delta y,\Delta z)$, dimensions $(w,l,h)$, heading $\gamma$, velocity $(v_x,v_y)$ and class identifier {\tt cls}.

During training, those $\bm{y}'_i$ with closest box parameters to ground truth are matched with the Hungarian algorithm in terms of $\ell_1$-distance, while all others are assigned a 'no-object' class id. An $\ell_1$-loss is then used to train the model. We use a pre-trained PointPillars backbone and refine it in combination with our model.

\subsection{Tracking}
A model overview of our proposed tracker is pictured in Fig.~\ref{fig_tracker_large}, (a) and (b).
To initialize the tracks, the detector as introduced above is run  on the first frame at timestep $t$, with queries $\bm{Y}_t=\{\bm{y}_{t,i}\}_{i=1}^M$, obtaining decoder output $\bm{y}'_{t,i}$ and (merely as a by-product) detections $\bm{d}_{t,i}$.
The high-dimensional vectors $\bm{y}'_{t,i}\in\mathbb{R}^d$, $i=1,\dots , M$ each contain \textit{encoded} information about an object, such as its dimension, location and object type as well as additional latent space. This makes them usable to carry tracking information over to the next timestep \cite{meinhardt_trackformer:_2021}. Therefore, we only use these $d$-dimensional vectors as tracker state, instead of the low-dimensional bounding boxes $\bm{d}_{t,i}$.

With lidar data, the ego vehicle can move a significant distance between two frames, making an ego-motion correction (EMC) necessary:
\begin{equation}\label{eq_EMC}
	\bm{y}''_{t,i}=\textrm{EMC}(\bm{y}'_{t,i}, \bm{\rho}_i, \bm{p}),
\end{equation}
where $\bm{p}$ is the ego pose change. Our EMC module is introduced in the following subsection.
The transformed output tokens $\bm{y}''_{t,k}$, $k=1,\dots,K$ that belong to an object are used as input to the transformer decoder in the next timestep, denoted as track queries, where  $K$ is the number of detected objects in frame $t$. Since it is always possible for a new track to spawn, the object queries $\bm{y}_{t+1,i}$, computed as in Eq.~\ref{eq_query_encoding} from newly sampled anchor locations $\bm{\rho}_{t+1,i}$, are passed into the decoder as before to serve as slots for new objects, in addition to the track queries. The set of decoder tokens is therefore $\bm{Y}_{t+1}=\{\bm{y}''_{t,k}\}_{k=1}^K\cup\{\bm{y}_{t+1,i}\}_{i=1}^M$. Due to the transformer design \cite{vaswani_attention_2017}, the number of tokens that is passed into the decoder at a time, i.e. track and object queries, does not impede with its performance: An additional token merely adds an additional participant in attention and therefore an additional output token.  

During training, pairs of consecutive frames are selected, whereas up to $n$ intermediate frames can be skipped, $n\in\mathbb{N}$. For the first frame of the pair, detections and corresponding decoder output $\bm{y}_i$ are computed. Those that were assigned closest to ground truth objects are then used as track queries for the second frame. From ground truth it is known which tracks are to be continued, which are occluded or terminated and which will spawn newly. Those that shall continue are assigned to their respective track queries' slots. Those estimates belonging to track queries whose track is occluded or terminated are assigned to the 'no-object' class. Ground truth objects belonging to new tracks on the other hand are assigned to the closest estimate that stems from an object query using the Hungarian algorithm. Like for the detector training, the $\ell_1$-loss is applied on the difference between the estimated box parameters and the ground truth.

At inference time, a track is considered to be continued if its corresponding track query results in a detection in the next frame of a confidence above a certain threshold $\lambda_{\textrm{track}}$. If it however results in a detection that is assigned to the 'no-object' class, the track is either terminated or occluded. To allow for recognition after occlusion, such track queries are carried over for some frames until the track is finally either terminated or continued. A new track on the other hand is spawned when a new object is detected with one of the object queries, which is above a certain confidence threshold $\lambda_{\textrm{detect}}$. We found that a non-maximum suppression, which favours boxes that stem from track queries over those originating from object queries, improves the tracking results.

We utilize track augmentation methods, following \cite{meinhardt_trackformer:_2021}: Since the event that new tracks spawn occurs rather rarely, it is simulated by removing each track query with a certain probability. A track query can also randomly spawn an additional, previously discarded, track query, which is meant to simulate occluded or terminated tracks.

\begin{figure}[!tbp]
	\centering
	\subfloat[Layer 0.]{\includegraphics[width=0.47\columnwidth]{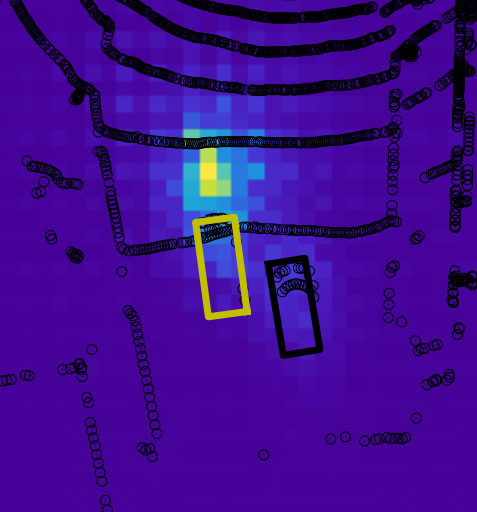}\label{fig:f1}}
	\hfill
	\subfloat[Layer 5.]{\includegraphics[width=0.47\columnwidth]{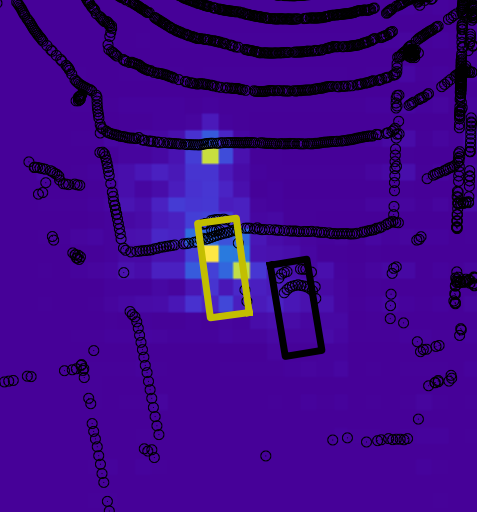}\label{fig:f2}}
	\caption{Attention visualization between one track query and each input grid cell. The tracked vehicle (yellow box) moved between the previous and this frame. The birds-eye-view grid cells are colored according to their attention score with the tracked vehicle's query, where yellow denotes larger values, and blue small ones. An attention shift is visible: In layer $0$, the track query seemingly expects the vehicle in the 'wrong' place, but it is able to correct its focus until the last decoder layer. In black, the lidar data of the current frame is visible, as well as a second vehicle nearby (for which the attention is not pictured).}
	\label{fig_AttnViz}
\end{figure}

In Fig.~\ref{fig_AttnViz}, the cross-attention in the decoder between one track query and the input birds-eye-view grid is visualized, where dark purple denotes low, and brighter colors high attention. The object that belongs to this track query (yellow) moved between the previous and this frame. This is why, in the first decoder layer, the query seems to search for its object in the 'wrong' place, but, due to its broad field of view in attention, it is able to shift its attention until the last decoder layer. Note that we do not explicitly model the motion of tracked agents, which might be a beneficial addition to the model in the future. We do, however, correct the motion of the ego-vehicle between frames, since this ego-motion would require too large attention shifts to be corrected by the transformer. The proposed EMC module for this is introduced in the following subsection.
\subsection{Ego-Motion Compensation}
 \begin{figure}[t]
	\centering
	\vspace{7pt}
	\includegraphics[trim={0.17cm 0 0 0},clip,width=\columnwidth]{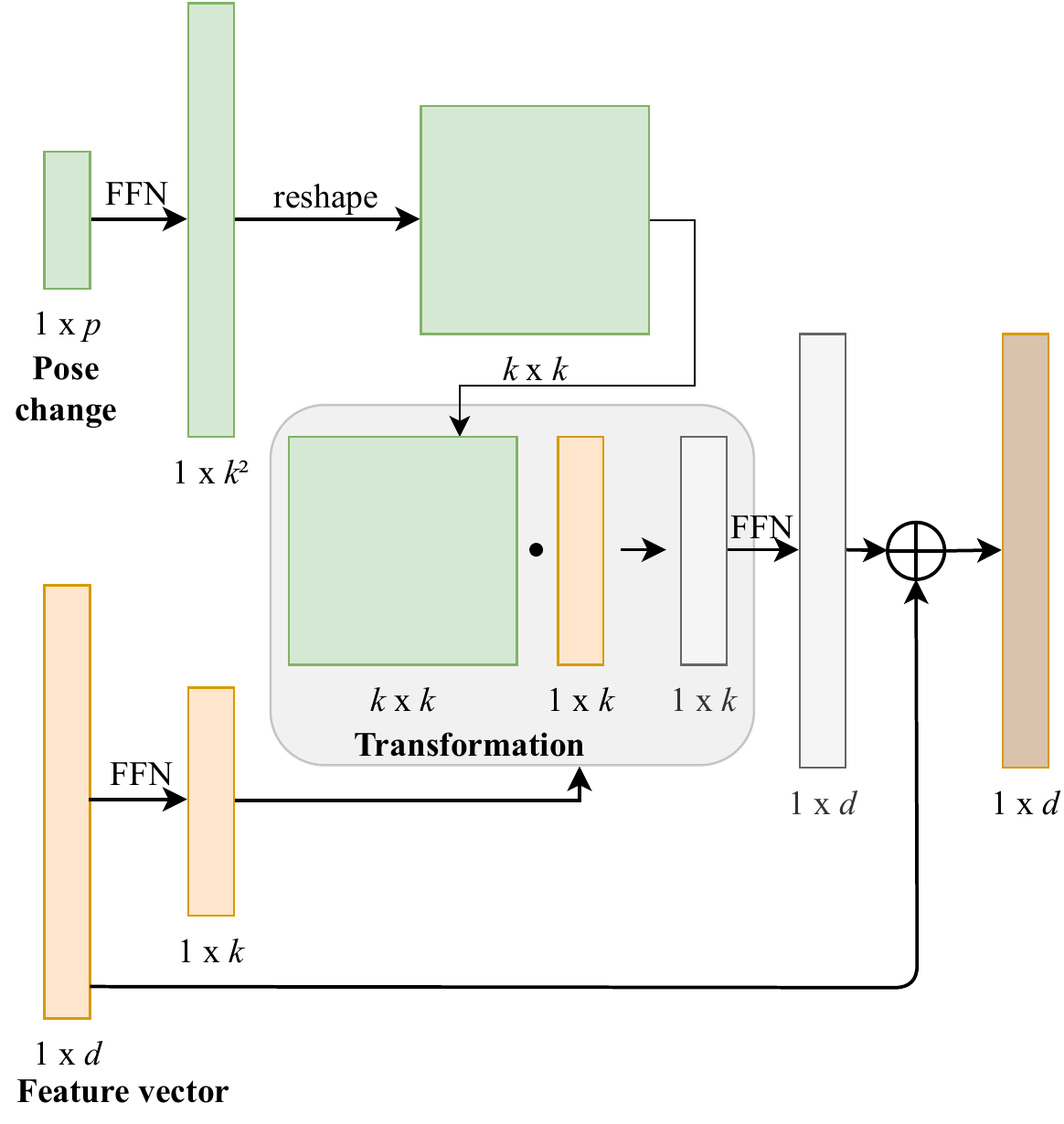}
	
	\caption{Ego-motion compensation (EMC) module overview. The ego-pose change consists of a translation and a rotation and it is converted into a transformation matrix through a feed-forward network. The feature vector that is to be ego-motion corrected is brought to the same length and the transformation is applied with a dot product, similar to a transformation in 3D space. A residual connection between the transformed and original feature vector is intended to prevent loss of information during the ego-motion correction.}
	\label{figEMC_module}
\end{figure}
An ego-vehicle can travel a significant distance between two lidar frames, making an ego-motion correction necessary. Since the tracker state is encoded in feature vectors rather than explicit parameters, such as bounding boxes, its transformation with the known pose change is non-trivial. Note that image-based trackers, such as \cite{meinhardt_trackformer:_2021}, do not require such a correction, since a moving ego-vehicle does not impact the position of objects in image space as much as it does in lidar space.

An alternative to correcting the internal state of the tracker may be to transform the input data into a certain global coordinate frame. However, after processing many consecutive frames, the area occupied by the measured data will become very large. This is problematic since the positional encoding at the transformer input would need to have an arbitrarily high resolution to cover the space.

Therefore, the only option for ego-motion compensation with our combined tracker and detector is shifting the internal tracker state according to the pose change. To achieve this task, we develop a novel ego-motion compensation (EMC) module as introduced in Eq.~\ref{eq_EMC}, which is pictured in Fig.~\ref{figEMC_module}. Our model is inspired by a transformation in 3D space: We compute a feature-agnostic transformation matrix $\bm{T}\in\mathbb{R}^{k\times k}$ from the known pose change $\bm{p}$ by using a feed-forward network (FFN) and reshaping the output, where $\bm{p}=(t_1,t_2,t_3, q_0,q_1,q_2,q_3)$ contains a translation as well as a rotation in quaternion form. With a second FFN, the feature vectors $\bm{y}'_i$ to be transformed are reduced to length $k$, after which the transformation takes place with a dot product. The result of the transformation is scaled back up to length $d$ with a FFN and a residual connection is added that is meant to prevent information loss. The overall model design serves two purposes: Firstly, the transformation is performed in a lower dimension than the full feature length to encourage only transforming necessary parts of the vector. Besides, $\bm{T}$ is computed without access to the feature vectors, aiming to make it independent of the location of the objects to be transformed.

For realistic training of the EMC module, nuScenes \cite{caesar_nuscenes:_2020} data is used. Each frame is first input into the transformer detector, which has previously been trained and is now used with fixed weights, to obtain the decoder output $\bm{y}'_i$. These vectors are input into the EMC module, together with the pose change $\bm{p}$ w.r.t. the following frame that is obtained from nuScenes meta data. To obtain ground truth data for training, the detector's FFN is used to compute box parameters $\bm{d}_i$ as defined in Eq.~\ref{eq_box_params} from the vectors $\bm{y}'_i$. With the pose change information, the expected location, heading and velocity of each box after transformation for the following frame is computed, using standard transformations in 3D space, resulting in parameters $(x',y',z'), \gamma', (v'_x,v'_y)$ for each object. Meanwhile, the size and class identifier shall be unchanged. Now, the change in location is implemented into a new, transformed anchor $\bm{\rho}'_i=(x',y',z')$, while the other parameters serve as ground truth for the EMC module:
\begin{equation}
\bm{d}'_i = (0, 0, 0,w,l,h,\gamma', v'_x,v'_y, {\tt cls}),
\end{equation}
where $\Delta x,\Delta y$ and $\Delta z$ are set to zero to align them with the new anchor location $\bm{\rho}'_i$.
To allow for large pose changes and to be prepared for possible frame skips in practice, the frame pairs used for training can have up to $n$ discarded frames between them, $n \in \mathbb{N}$.
\vspace{5pt}
\section{RESULTS}
We evaluate each component of the proposed model, i.e. the EMC module, the object detector and the tracker. For object detection and tracking, the nuScenes metrics, as defined in \cite{caesar_nuscenes:_2020}, are used.

We show exemplary results for the class 'car' in this section. It is straightforward to incorporate many different classes as described in \cite{carion_end--end_2020}. The required number of queries and, thus, the memory requirement of the model does, however, depend linearly on the \textit{maximum} number of expected objects throughout all frames. Since the number of objects fluctuates heavily between frames in the nuScenes dataset, it is a current topic of research to adapt the number of queries dynamically based on an expected number of objects in the current frame.

\subsection{Ego-Motion Compensation}
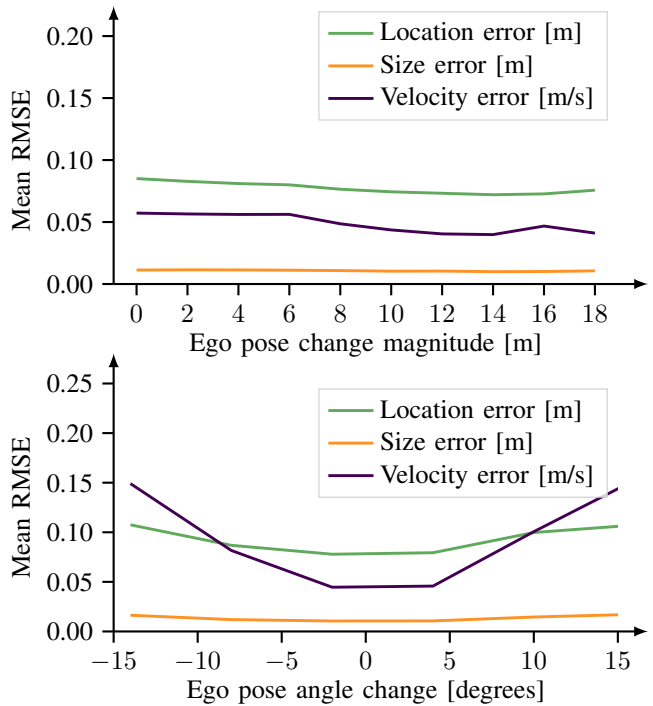
\begin{figure}[t]
	\vspace{7pt}
\begin{tikzpicture}

\definecolor{color0}{rgb}{0.3803921568627451, 0.6627450980392157, 0.3607843137254902}
\definecolor{color1}{rgb}{1.0, 0.5725490196078431, 0.1411764705882353}
\definecolor{color2}{rgb}{0.26666666666666666, 0.00392156862745098, 0.32941176470588235}
\begin{axis}[
height=138.86214363138544, 
legend cell align={left},
legend style={at={(0.98,1.1)},fill opacity=0.8, draw opacity=1, text opacity=1, draw=white!80!black},
tick align=outside,
tick pos=left,
width=235.71811,
x grid style={white!69.0196078431373!black},
xlabel={Ego pose change magnitude [m]},
xmin=-0.9, xmax=18.9,
xtick style={color=black},
y grid style={white!69.0196078431373!black},
ylabel={Mean RMSE},
ymin=0, ymax=0.2,
ytick style={color=black},
ytick={0,0.05,0.1,0.15,0.2,0.25,0.3},
yticklabels={0.00,0.05,0.10,0.15,0.20,0.25,0.30}
]
\addplot [semithick, color0]
table {%
	0 0.0850383639335632
	2 0.0828097462654114
	4 0.0810361132025719
	6 0.0799940675497055
	8 0.0764643028378487
	10 0.0743804797530174
	12 0.0732044652104378
	14 0.0720195174217224
	16 0.0726975798606873
	18 0.0756773352622986
};
\addlegendentry{Location error [m]}
\addplot [semithick, color1]
table {%
0 0.0112565467134118
2 0.0114517454057932
4 0.0113723380491138
6 0.0111448187381029
8 0.0108375819399953
10 0.0103735653683543
12 0.0104458155110478
14 0.00996312871575356
16 0.0100831324234605
18 0.0106163900345564
};
\addlegendentry{Size error [m]}

\addplot [semithick, color2]
table {%
	0 0.057215828448534
	2 0.0565180480480194
	4 0.0561173111200333
	6 0.0561839155852795
	8 0.0485810786485672
	10 0.0436305440962315
	12 0.0404257886111736
	14 0.0398746505379677
	16 0.0467552095651627
	18 0.0410616509616375
};
\addlegendentry{Velocity error [m/s]}
\end{axis}

\end{tikzpicture}
\begin{tikzpicture}

\definecolor{color0}{rgb}{0.3803921568627451, 0.6627450980392157, 0.3607843137254902}
\definecolor{color1}{rgb}{1.0, 0.5725490196078431, 0.1411764705882353}
\definecolor{color2}{rgb}{0.26666666666666666, 0.00392156862745098, 0.32941176470588235}
\begin{axis}[
height=138.86214363138544,
legend cell align={left},
legend style={fill opacity=0.8, draw opacity=1, text opacity=1, draw=white!80!black},
tick align=outside,
tick pos=left,
width=235.71811,
x grid style={white!69.0196078431373!black},
xlabel={Ego pose angle change [degrees]},
xmin=-15, xmax=15,
xtick style={color=black},
y grid style={white!69.0196078431373!black},
ylabel={Mean RMSE},
ymin=0, ymax=0.25,
ytick style={color=black},
ytick={0,0.05,0.1,0.15,0.2,0.25,0.3},
yticklabels={0.00,0.05,0.10,0.15,0.20,0.25,0.30}
]
\addplot [semithick, color0]
table {%
-14 0.107494071125984
-8 0.0868793353438377
-2 0.0778678581118584
4 0.0793842524290085
10 0.0997385606169701
16 0.107244364917278
};
\addlegendentry{Location error [m]}
\addplot [semithick, color1]
table {%
-14 0.0163502283394337
-8 0.0119561655446887
-2 0.010543018579483
4 0.0106332246214151
10 0.0145654361695051
16 0.0172499213367701
};
\addlegendentry{Size error [m]}

\addplot [semithick, color2]
table {%
-14 0.149072870612144
-8 0.0816687718033791
-2 0.044602807611227
4 0.0457384437322617
10 0.100617781281471
16 0.152506485581398
};
\addlegendentry{Velocity error [m/s]}
\end{axis}

\end{tikzpicture}
	\caption{EMC module evaluation on frame pairs from nuScenes {\tt val} dataset. The frame pairs were either be one, two or three frame lengths apart.}
	\label{figEMC_rmse2}
\end{figure}
The EMC module evaluation is performed on the nuScenes {\tt val} dataset. For this, a frame is first drawn, and detections $\bm{d}_i$ as well as the corresponding feature vectors $\bm{y}'_i$ are computed using the fully trained transformer detector. A second frame is loaded, either one, two or three frame lengths~($\Delta t$) apart from the first. The feature vectors and the ego pose change $\bm{p}$ between the two frames are input into the EMC module obtaining the results $\bm{y}''_i$. To generate ground truth, the box parameters $\bm{d}_i$ are ego-motion corrected manually as described before and compared to the box parameters that result from the EMC module's output through the estimation head, $\textrm{FFN}(\bm{y}''_i)$.

To gain an understanding whether the error depends on the extent of the ego pose change, mean RMSE values are plotted in terms of the absolute length of the translation in the pose change, see Fig.~\ref{figEMC_rmse2} (top). We find that the error does not depend on the pose change magnitude, which can be explained with the fact that a similar distribution of pose change magnitudes was present during EMC training. This means that the EMC module has no issue transforming the track queries to the next available frame, even if two frame skips occurred in between. The RMSE for size is generally smaller than for location and velocity, because it is supposed to be unchanged during ego-motion correction, which the model seems to be able to obey.

Concerning the RMSE in terms of the angle in the rotation component of the ego pose change, see Fig.~\ref{figEMC_rmse2} (bottom). We find that $95\%$ of pose change angles in nuScenes data (with up to two frame skips in between) lay in the range of $-6$ to $6$ degrees. Therefore, the EMC module performs best in this range, while it is still able to correct larger ego pose rotation angles. If another distribution of angles is desired, the pose change parameters can be curated manually during training, rather than using those from nuScenes metadata.

\subsection{Object Detection}
The nuScenes metrics \cite{caesar_nuscenes:_2020} for detection of cars on the nuScenes {\tt val} dataset can be found in Table~\ref{tab:object_detection}. We find that the detector part of our model can not reach state of the art results yet. However, it does outperform its PointPillars \cite{lang_pointpillars:_2019} basis. More research is necessary on detection on lidar data with transformers which will directly benefit our model. A more powerful backbone may also improve performance.
\begin{table}[h]
	\setlength\tabcolsep{0pt} 
	\caption{Object detection results (car)}
	\label{tab:object_detection}
	\begin{tabular*}{\columnwidth}{@{\extracolsep{\fill}} ll cccccc}
		\toprule
		Method  & AP$\uparrow$ & ATE (m)$\downarrow$& ASE (1-IOU) $\downarrow$& AOE (rad)$\downarrow$\\
		\midrule
		PointPillars \cite{lang_pointpillars:_2019} &0.684 &
		\textbf{0.281}& 0.164&0.204\\
		Transformer det. (ours)   &\textbf{0.727}&	0.284&	\textbf{0.161}&\textbf{0.076} \\
		\bottomrule
		
	\end{tabular*}
\end{table}
\subsection{Tracking}
We evaluate the performance of our tracking model on the nuScenes {\tt val} dataset. The nuScenes metrics \cite{caesar_nuscenes:_2020} for the category car can be found in Table~\ref{tab:nuscenes_metrics}. TransMOT was trained for 300 epochs on an NVIDIA A100 GPU on the nuScenes {\tt train} dataset with a learning rate of $3\cdot 10^{-5}$, $n=2$ maximum intermediate frame skips and $M=100$ object queries. During inference, the confidence thresholds are set to $\lambda_{\textrm{detect}}=0.95$, $\lambda_{\textrm{track}}=0.5$.

Our results are compared with the ProbMOT \cite{chiu_probabilistic_2020} baseline that is based on a Kalman filter as described in section~\ref{sec_related_work_MOT}. We choose this comparison in order to investigate the model's tracking capabilities independent from its object detection capability. For this, bounding boxes as output by our transformer detector module are passed to the ProbMOT model for tracking. The default parameters of the model for nuScenes are used \cite{chiu_probabilistic_2020}. In the table, \textit{thr} denotes the confidence threshold above which boxes were given to the tracker.
Although our model follows a radically new approach, it is able to outperform the ProbMOT baseline. This shows the potential of our proposed method, since it can still be extended, e.g. with a better detector, or a motion model per tracked object. Such improvements will enable it to further outperform its traditional Bayesian baseline.

As mentioned in the introduction, skipped frames or delayed sensor input is a major issue in practice. To evaluate the susceptibility to such events, we run our tracking model on all scenes of the {\tt val} dataset. Each frame has a certain probability $p$ to be dropped and withheld from the tracker. The results are pictured in Figure~\ref{figFrameSkips}. Interestingly, TransMOT can handle many frame skips without a noticeable performance drop, even multiple skips at once are not an issue. This makes the tracker robust for real-world scenarios. It outperforms ProbMOT clearly in this regard, although Kalman filter based methods are known for their ability to predict the per-agent movement. 
\begin{figure}[t]
	\vspace{15pt}
\begin{tikzpicture}

\definecolor{color0}{rgb}{0.3803921568627451, 0.6627450980392157, 0.3607843137254902}
\definecolor{color1}{rgb}{1.0, 0.5725490196078431, 0.1411764705882353}

\begin{axis}[
height=138.86214363138544,
legend cell align={left},
legend style={at={(0.04,0.06)},anchor=south west,fill opacity=0.8, draw opacity=1, text opacity=1, draw=white!80!black},
tick align=outside,
tick pos=left,
width=235.71811,
x grid style={white!69.0196078431373!black},
xlabel={Frame skip probability $p$},
xmin=0, 
xtick style={color=black},
y grid style={white!69.0196078431373!black},
ylabel={AMOTA},
ymin=0.3, 
ytick style={color=black},
]
\addplot [semithick, color0]
table {%
0 0.674
0.1 0.672
0.2 0.669
0.3 0.665
0.4 0.661
0.5 0.635
0.6 0.602
0.7 0.562
};
\addlegendentry{TransMOT (ours)}
\addplot [semithick, color1]
table {%
0 0.65
0.1 0.621
0.2 0.585
0.3 0.566
0.4 0.523
0.5 0.483
0.6 0.418
0.7 0.374
};
\addlegendentry{ProbMOT \cite{chiu_probabilistic_2020}}
\end{axis}

\end{tikzpicture}
	\caption{Analysis of susceptibility to skipped frames. Each frame is withheld from the model with probability $p$. Our model is only susceptible to very large frame skip probabilities.}
	\label{figFrameSkips}
\end{figure}
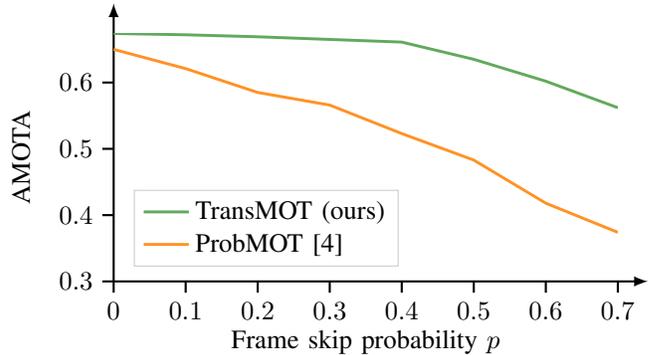
\begin{table}[h]
	\setlength\tabcolsep{0pt} 
	\caption{Tracking performance on transformer detections (car)}
	\label{tab:nuscenes_metrics}
	\begin{tabular*}{\columnwidth}{@{\extracolsep{\fill}} lc cccccc}
		\toprule
		Method & thr & AMOTA$\uparrow$ & AMOTP$\downarrow$& MT$\uparrow$ & FP$\downarrow$ & FN$\downarrow$ & IDS$\downarrow$\\
		\midrule
		\multirow{4}{*}{ProbMOT \cite{chiu_probabilistic_2020}}& 0.3 & 0.596&	0.880&	1497 & 8018&	19853&	1677 \\
		& 0.5 & 0.645&	0.828&	1721 & 6505&	16801&	542 \\
		&0.7 & 0.650 &  0.822 &  1689 &  6297  &  16809 &  559 \\
		&0.9 &0.614&	0.897&	1522 & \textbf{5167}&	18216&	\textbf{502}	 \\
		\addlinespace
		TransMOT (ours)  & - &\textbf{0.674}&	\textbf{0.754}&	\textbf{2096} & 9449&	\textbf{14071}&	1403 \\
		\bottomrule
	\end{tabular*}
\end{table}

\subsection{Ablation Study}
To evaluate the necessity of our ego-motion compensation (EMC) module, we perform an ablation study.
First, a tracker model is trained that has no ego-motion compensation, while all other parameters remain unchanged. For this, the feature vectors belonging to found objects are used as track queries in the next timestep without modifying them at all. This change leads to a significant performance drop in terms of the average multi-object tracking accuracy (AMOTA \cite{caesar_nuscenes:_2020}), as noted in Table~\ref{tab:ablation_emc}. Another experiment we perform is to only transform the anchor locations that belong to the track queries, but to not correct the other parameters. We find that this also leads to a performance drop of the model.
\begin{table}[h]
	\setlength\tabcolsep{0pt} 
	\caption{Tracking with and without EMC}
	\label{tab:ablation_emc}
	\begin{tabular*}{\columnwidth}{@{\extracolsep{\fill}} ll cccccc}
		\toprule
		Method  & AMOTA$\uparrow$ & AMOTP$\downarrow$& MT$\uparrow$ & FP$\downarrow$ & FN$\downarrow$ & IDS$\downarrow$\\
		\midrule
		Without EMC   &0.564&	0.994&	1703 & \textbf{8364}	&17162&	4829 \\
		Anchor transform   &0.657&	0.791&	2020 & 9307&	15070&	1699 \\
		\addlinespace
		With EMC   &\textbf{0.674}&	\textbf{0.754}&	\textbf{2096} & 9449&	\textbf{14071}&	\textbf{1403} \\
		\bottomrule
	\end{tabular*}
\end{table}
\addtolength{\textheight}{-4.1cm}   
\section{CONCLUSION AND OUTLOOK}
We presented TransMOT, a novel joint detection-and-tracking approach for point cloud data based on a transformer. We utilize the detector's latent space as tracker state and are able to outperform a Kalman filter-based baseline method on our given detections. We believe that the future of tracking will be in such joint models.

The detector component of TransMOT is able to outperform its PointPillars baseline. Besides this, the current surge in research on computer vision with transformers can contribute to improve its performance in the future, which, in turn, will improve our tracker. One example for ongoing research in the area of transformer-based object detection is the very recent publication \cite{bai2022transfusion}, which reaches state of the art results for lidar-based object detection and can be used as a building block within our proposed tracking model.

TransMOT is applicable to automotive lidar data, which was not possible with prior transformer-based trackers. It is able to handle sensor input at arbitrary timesteps and includes a novel ego-motion compensation module. In addition, it is very modular and can be extended in numerous ways: For example, the extent of the EMC module can be increased to also include a per-agent motion compensation, similar to the motion model in the prediction step of a Kalman filter. It is also possible to feed an additional data source into a transformer \cite{hu_transformer_2021}, such as camera data.




\bibliographystyle{IEEEtran}

\bibliography{IEEEabrv,cleaned_references_long}

\begin{thebibliography}{10}
\providecommand{\url}[1]{#1}
\csname url@samestyle\endcsname
\providecommand{\newblock}{\relax}
\providecommand{\bibinfo}[2]{#2}
\providecommand{\BIBentrySTDinterwordspacing}{\spaceskip=0pt\relax}
\providecommand{\BIBentryALTinterwordstretchfactor}{4}
\providecommand{\BIBentryALTinterwordspacing}{\spaceskip=\fontdimen2\font plus
\BIBentryALTinterwordstretchfactor\fontdimen3\font minus
  \fontdimen4\font\relax}
\providecommand{\BIBforeignlanguage}[2]{{%
\expandafter\ifx\csname l@#1\endcsname\relax
\typeout{** WARNING: IEEEtran.bst: No hyphenation pattern has been}%
\typeout{** loaded for the language `#1'. Using the pattern for}%
\typeout{** the default language instead.}%
\else
\language=\csname l@#1\endcsname
\fi
#2}}
\providecommand{\BIBdecl}{\relax}
\BIBdecl

\bibitem{bar2011tracking}
Y.~Bar-Shalom, P.~K. Willett, and X.~Tian, \emph{{Tracking and Data Fusion: A
  Handbook of Algorithms}}.\hskip 1em plus 0.5em minus 0.4em\relax YBS
  Publishing, CT, USA, 2011, vol.~11.

\bibitem{mahler2007statistical}
R.~P. Mahler, \emph{{Statistical Multisource-Multitarget Information
  Fusion}}.\hskip 1em plus 0.5em minus 0.4em\relax Artech House Norwood, MA,
  USA, 2007.

\bibitem{weng_3d_2019}
X.~Weng, J.~Wang, D.~Held, and K.~Kitani, ``{3D Multi-Object Tracking: A
  Baseline and New Evaluation Metrics},'' in \emph{{2020 IEEE/RSJ International
  Conference on Intelligent Robots and Systems (IROS)}}, 2020, pp.
  10\,359--10\,366.

\bibitem{chiu_probabilistic_2020}
H.-K. Chiu, A.~Prioletti, J.~Li, and J.~Bohg, ``Probabilistic 3{D}
  {Multi}-{Object} {Tracking} for {Autonomous} {Driving},'' \emph{arXiv
  preprint arXiv:2001.05673}, 2020.

\bibitem{caesar_nuscenes:_2020}
H.~Caesar, V.~Bankiti, A.~H. Lang, S.~Vora, V.~E. Liong, Q.~Xu, A.~Krishnan,
  Y.~Pan, G.~Baldan, and O.~Beijbom, ``{nuScenes: {A} Multimodal Dataset for
  Autonomous Driving},'' in \emph{Proceedings of the IEEE conference on
  Computer Vision and Pattern Recognition}, 2020, pp. 11\,621--11\,631.

\bibitem{zhang_robust_2019}
W.~Zhang, H.~Zhou, S.~Sun, Z.~Wang, J.~Shi, and C.~C. Loy, ``{Robust
  Multi-Modality Multi-Object Tracking},'' in \emph{Proceedings of the IEEE/CVF
  International Conference on Computer Vision}, 2019, pp. 2365--2374.

\bibitem{chiu_probabilistic_2020-1}
H.-K. Chiu, J.~Li, R.~Ambruş, and J.~Bohg, ``{Probabilistic 3D Multi-Modal,
  Multi-Object Tracking for Autonomous Driving},'' in \emph{{2021 IEEE
  International Conference on Robotics and Automation (ICRA) }}, 2021, pp.
  14\,227--14\,233.

\bibitem{meinhardt_trackformer:_2021}
T.~Meinhardt, A.~Kirillov, L.~Leal-Taixe, and C.~Feichtenhofer,
  ``{TrackFormer}: {Multi}-{Object} {Tracking} with {Transformers},''
  \emph{arXiv preprint arXiv:2101.02702}, 2021.

\bibitem{luo_fast_2018}
W.~Luo, B.~Yang, and R.~Urtasun, ``{Fast and furious: {Real} Time End-to-End 3D
  Detection, Tracking and Motion Forecasting},'' in \emph{Proceedings of the
  IEEE conference on Computer Vision and Pattern Recognition}, 2018, pp.
  3569--3577.

\bibitem{vaswani_attention_2017}
A.~Vaswani, N.~Shazeer, N.~Parmar, J.~Uszkoreit, L.~Jones, A.~N. Gomez, L.~u.
  Kaiser, and I.~Polosukhin, ``{Attention is All You Need},'' in
  \emph{{Advances in Neural Information Processing Systems}}, 2017, pp.
  5998--6008.

\bibitem{dosovitskiy_image_2020}
A.~Dosovitskiy, L.~Beyer, A.~Kolesnikov, D.~Weissenborn, X.~Zhai,
  T.~Unterthiner, M.~Dehghani, M.~Minderer, G.~Heigold, S.~Gelly \emph{et~al.},
  ``{An Image is Worth 16x16 Words: Transformers for Image Recognition at
  Scale},'' in \emph{International Conference on Learning Representations},
  2021.

\bibitem{carion_end--end_2020}
N.~Carion, F.~Massa, G.~Synnaeve, N.~Usunier, A.~Kirillov, and S.~Zagoruyko,
  ``{End-to-End Object Detection with Transformers},'' in \emph{{European
  conference on computer vision}}.\hskip 1em plus 0.5em minus 0.4em\relax
  Springer, 2020, pp. 213--229.

\bibitem{engel_point_2020}
N.~Engel, V.~Belagiannis, and K.~Dietmayer, ``{Point Transformer},'' \emph{IEEE
  Access}, vol.~9, pp. 134\,826--134\,840, 2021.

\bibitem{shin_roarnet_2019}
K.~Shin, Y.~P. Kwon, and M.~Tomizuka, ``{RoarNet: A Robust 3D Object Detection
  based on RegiOn Approximation Refinement},'' in \emph{2019 IEEE Intelligent
  Vehicles Symposium (IV)}, 2019, pp. 2510--2515.

\bibitem{ciaparrone2020deep}
G.~Ciaparrone, F.~L. S{\'a}nchez, S.~Tabik, L.~Troiano, R.~Tagliaferri, and
  F.~Herrera, ``{Deep Learning in Video Multi-object tracking: A Survey},''
  \emph{Neurocomputing}, vol. 381, pp. 61--88, 2020.

\bibitem{zhou2020tracking}
X.~Zhou, V.~Koltun, and P.~Kr{\"a}henb{\"u}hl, ``{Tracking Objects as
  Points},'' in \emph{{European conference on computer vision}}.\hskip 1em plus
  0.5em minus 0.4em\relax Springer, 2020, pp. 474--490.

\bibitem{bergmann2019tracking}
P.~Bergmann, T.~Meinhardt, and L.~Leal-Taixe, ``{Tracking without bells and
  whistles},'' in \emph{Proceedings of the IEEE conference on Computer Vision
  and Pattern Recognition}, 2019, pp. 941--951.

\bibitem{xu2020train}
Y.~Xu, A.~Osep, Y.~Ban, R.~Horaud, L.~Leal-Taix{\'e}, and X.~Alameda-Pineda,
  ``{How to Train Your Deep Multi-object Tracker},'' in \emph{Proceedings of
  the IEEE conference on Computer Vision and Pattern Recognition}, 2020, pp.
  6787--6796.

\bibitem{fortin2015model}
B.~Fortin, R.~Lherbier, and J.-C. Noyer, ``{A Model-based Joint Detection and
  Tracking Approach for Multi-Vehicle Tracking with Lidar Sensor},''
  \emph{{IEEE Transactions on Intelligent Transportation Systems}}, vol.~16,
  no.~4, pp. 1883--1895, 2015.

\bibitem{huang2021joint}
K.~Huang and Q.~Hao, ``{Joint Multi-Object Detection and Tracking with
  Camera-LiDAR Fusion for Autonomous Driving},'' in \emph{2021 IEEE/RSJ
  International Conference on Intelligent Robots and Systems (IROS)}.\hskip 1em
  plus 0.5em minus 0.4em\relax IEEE, 2021, pp. 6983--6989.

\bibitem{liang2020pnpnet}
M.~Liang, B.~Yang, W.~Zeng, Y.~Chen, R.~Hu, S.~Casas, and R.~Urtasun,
  ``{Pnpnet: End-to-end Perception and Prediction with Tracking in the Loop},''
  in \emph{Proceedings of the IEEE conference on Computer Vision and Pattern
  Recognition}, 2020, pp. 11\,553--11\,562.

\bibitem{lang_pointpillars:_2019}
A.~H. Lang, S.~Vora, H.~Caesar, L.~Zhou, J.~Yang, and O.~Beijbom,
  ``{Pointpillars: {Fast} Encoders for Object Detection from Point Clouds},''
  in \emph{Proceedings of the IEEE conference on Computer Vision and Pattern
  Recognition}, 2019, pp. 12\,697--12\,705.

\bibitem{qi_pointnet++_2017}
C.~R. Qi, L.~Yi, H.~Su, and L.~J. Guibas, ``{PointNet++: Deep Hierarchical
  Feature Learning on Point Sets in a Metric Space},'' \emph{Advances in Neural
  Information Processing Systems}, vol.~30, 2017.

\bibitem{wang_linformer:_2020}
S.~Wang, B.~Z. Li, M.~Khabsa, H.~Fang, and H.~Ma, ``Linformer:
  {Self}-{Attention} with {Linear} {Complexity},'' \emph{arXiv preprint
  arXiv:2006.04768}, 2020.

\bibitem{tay_efficient_2020}
Y.~Tay, M.~Dehghani, D.~Bahri, and D.~Metzler, ``Efficient {Transformers}: {A}
  {Survey},'' \emph{arXiv preprint arXiv:2009.06732}, 2020.

\bibitem{misra_end--end_2021}
I.~Misra, R.~Girdhar, and A.~Joulin, ``{An End-to-End Transformer Model for 3D
  Object Detection},'' in \emph{{Proceedings of the IEEE/CVF International
  Conference on Computer Vision}}, 2021, pp. 2906--2917.

\bibitem{tancik_fourier_2020}
M.~Tancik, P.~Srinivasan, B.~Mildenhall, S.~Fridovich-Keil, N.~Raghavan,
  U.~Singhal, R.~Ramamoorthi, J.~Barron, and R.~Ng, ``{Fourier features let
  networks learn high frequency functions in low dimensional domains},''
  \emph{arXiv preprint arXiv:2006.10739}, 2020.

\bibitem{rahimi2007random}
A.~Rahimi and B.~Recht, ``{Random features for large-scale kernel machines},''
  \emph{Advances in neural information processing systems}, vol.~20, 2007.

\bibitem{bai2022transfusion}
X.~Bai, Z.~Hu, X.~Zhu, Q.~Huang, Y.~Chen, H.~Fu, and C.-L. Tai, ``{TransFusion:
  Robust LiDAR-Camera Fusion for 3D Object Detection with Transformers},''
  \emph{arXiv preprint arXiv:2203.11496}, 2022.

\bibitem{hu_transformer_2021}
R.~Hu and A.~Singh, ``{UniT: Multimodal Multitask Learning with a Unified
  Transformer},'' \emph{arXiv preprint arXiv:2102.10772}, 2021.

\end{thebibliography}

\end{document}